# Semi-supervised learning with Bayesian Confidence Propagation Neural Network

Naresh Balaji Ravichandran[1], Anders Lansner[1,2], Pawel Herman[1] *

1- Computational Brain Science Lab, KTH Royal Institute of Technology
10044 Stockholm, Sweden
2- Department of Mathematics, Stockholm University
10691, Stockholm, Sweden

**Abstract.** Learning internal representations from data using no or few labels is useful for machine learning research, as it allows using massive amounts of unlabeled data. In this work, we use the Bayesian Confidence Propagation Neural Network (BCPNN) model developed as a biologically plausible model of the cortex. Recent work has demonstrated that these networks can learn useful internal representations from data using local Bayesian-Hebbian learning rules. In this work, we show how such representations can be leveraged in a semi-supervised setting by introducing and comparing different classifiers. We also evaluate and compare such networks with other popular semi-supervised classifiers.

## 1 Introduction

Artificial neural networks have made remarkable progress in supervised pattern recognition in recent years. In particular, deep neural networks have dominated the field largely due to their capability to discover hierarchies of useful internal (hidden) representations from data. However, most deep learning methods rely extensively on labelled samples for extracting such representations. Semi-supervised learning methods aim to remedy this problem by learning effectively from few labelled samples [1].

Neural network approaches to semi-supervised learning can be broadly categorized into two types. The first is predominantly a supervised approach that consists in treating unlabeled data as cases of "missing labels" and applying different regularizations to loss function to ensure label prediction consistency [2,3]. The second type, reflected in our work, involves unsupervised learning to build internal representations, followed by supervised learning with the available labelled samples [4]. These models typically learn end-to-end using backprop. Although backprop as a learning algorithm is highly efficient for finding good representations from data, it is an unlikely candidate model for synaptic plasticity in the brain. In this work, we show how neural networks built by following basic brain-like principles like incremental local Hebbian learning, divisive normalization, sparse connectivity, sparse activity, and modular architecture, can be used for semi-supervised learning.

* Funding for the work is received from the Swedish e-Science Research Centre (SeRC), Digital Futures, Vetenskapsrådet 2018-05360, European Commission H2020 program, Grant Agreement No. 800999 (SAGE2), and Grant Agreement No. 801039 (EPiGRAM-HS). The simulations were performed on resources provided by Swedish National Infrastructure for Computing (SNIC) at the PDC Center for High Performance Computing, KTH Royal Institute of Technology.

## 2 Model Description

### 2.1 Unsupervised learning of internal representations

We now describe the BCPNN architecture and update rules. BCPNN is a biologically plausible model that frames neural computation, i.e., synaptic learning and activity update rules, as probabilistic computations [5-7]. The network architecture is derived from the columnar organization of the cortex [8] and can be summarized as follows: each layer comprises $n_{HC}$ modules called hypercolumns, each of which in turn comprise $n_{MC}$ minicolumns (typically $n_{MC} = 100$). When a source layer activates a target hypercolumn, the update reflects the posterior probability of the target conditioned on the source layer (Fig. 1A). We will use $\pi$ to refer to the activities, and $X, Y$, and $Z$ to refer to the input, hidden, and output layer of the network, respectively. The activities are updated as follows:

$$s(y_j) = b(y_j) + \sum_{i=1}^{n_{HC}^{inp}} \sum_{x_i=1}^{n_{MC}^{inp}} \pi(x_i)\, w(x_i, y_j)$$
$$\pi(y_j) = \left. e^{s(y_j)} \middle/ \sum_{y_k=1}^{n_{MC}^{hid}} e^{s(y_k)} \right. \qquad (1)$$

where $s(y_j)$ is the total input received by the target minicolumn from which the activity $\pi(y_j)$ is recovered by softmax normalization within the hypercolumn. The terms $b$ and $w$ refer to the bias and weight. During learning, individual and joint (marginal) probabilities, biases, and weights are incrementally updated as follows:

$$\tau_p \frac{d}{dt} p(x_i) = \kappa\, (\pi(x_i) - p(x_i)),$$
$$\tau_p \frac{d}{dt} p(y_j) = \kappa\, (\pi(y_j) - p(y_j)),$$
$$\tau_p \frac{d}{dt} p(x_i, y_j) = \kappa\, (\pi(x_i)\, \pi(y_j) - p(x_i, y_j)),$$
$$b(y_j) = \log p(y_j),$$
$$w(x_i, y_j) = \log \frac{p(x_i, y_j)}{p(x_i) p(y_j)} \qquad (2)$$

where $\tau_p$ is the learning time constant, and $\kappa$ is the global reward signal. As a crucial departure from traditional deep networks, the learning rule is local, associative, and Hebbian [5,6], i.e., dependent only on the presynaptic and postsynaptic activities (apart from the global reward signal). Recent work [5,6] has shown that the BCPNN architecture can also learn internal representations in a fully unsupervised manner (Fig. 1B). This can be accomplished when coupled with a structural plasticity mechanism that learns a sparse set of connections. Such representations when evaluated with a linear classifier reach 98.6% on MNIST and 88.9% on Fashion-MNIST [6]. This performance is better than unsupervised machine learning models with a similar architecture like autoencoders, restricted Boltzmann machines, and on par with a simple supervised multi-layer perceptron trained with backprop [6].

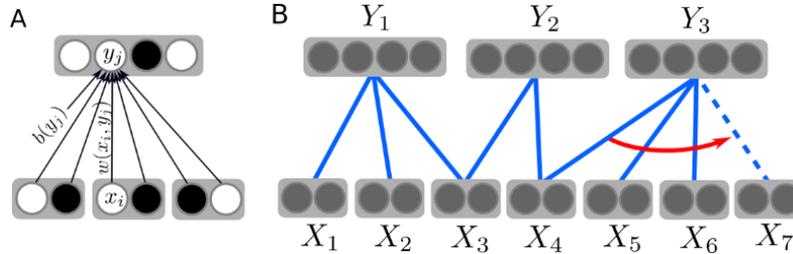

Fig. 1: Schematic of the BCPNN architecture for learning internal representations. (A) Network with three input hypercolumns (binary) and one hidden hypercolumn (multinomial) (gray boxes). (B) Structural plasticity allows learning distributed internal representation by forming a sparse set of connections (blue lines).

### 2.2 Semi-supervised learning with Associative and Go/No-go classifiers

Here we explain two different network models (Fig. 2) for classification of the internal representations. The first one is a simple associative classifier (Assoc., Fig. 2A) employing the BCPNN learning rule, as described in Section 3.1, with $\kappa = 1$ (Equation 2). In the learning process we first drive the hidden layer activities from the input layer for each labelled sample, and then clamp the labels in the output layer. The weights from the hidden to output layer are updated incrementally.

The second classifier is the Go/No-go model (Fig. 2B) inspired by the action selection model of the basal ganglia dual pathways organization [7], where the Go pathway promotes the correct choice, while the No-go pathway demotes the incorrect choice. Since No-go pathway has a negative effect on the output, we multiply the total input received by the minicolumns from the No-go connections by -1 before normalization. We evaluated this model using three separate networks, Go, No-go, and Go/No-go. The outputs minicolumns of Go/No-go network receive input from both Go and No-go connections. For training the Go connections, the correct label is clamped, and for the No-go connections the wrongly predicted label is clamped. Both pathways use the same BCPNN rule (Equations 1 and 2) with the global factor $\kappa$ set to the prediction error signal:

$$\kappa_{Go} = 1 - \pi_{corr}, \qquad \kappa_{No-go} = \begin{cases} 0, & if\ corr = pred \\ \pi_{pred}, & otherwise \end{cases} \qquad (3)$$

where $\pi$ is the activity in the output layer (shorthand for $\pi(z)$), $corr$ and $pred$ refer to the label index and the output index with the highest prediction, respectively. The learning converges with both $\kappa$'s reaching zero when all the correct labels are selected by the Go pathway and the incorrect labels are cancelled by the No-go pathway.

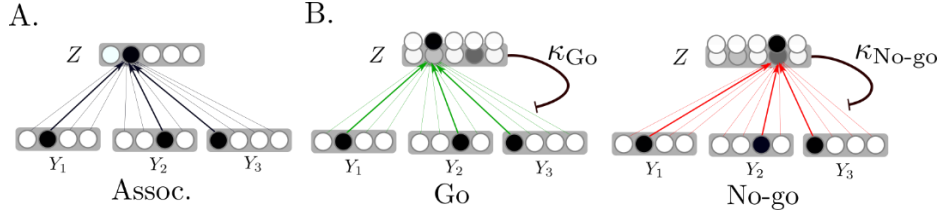

Fig. 2: Illustration of supervised learning update for one labelled sample. (A) Assoc. classifier learns weights between the hidden activations and the correct label activations ($corr = 2$ in the illustration). (B) Go classifier learns the correct label, while No-go classifier learns the wrongly predicted label ($pred = 4$). Suppose $\pi_{corr} = 0.3$ and $\pi_{pred} = 0.6$ (the remaining probability of 0.1 is assigned to other outputs), the Go pathway has $\kappa_{Go} = 0.7$ and the No-go pathway has $\kappa_{No-go} = 0.6$ (width of the connections reflect the weight update for one labelled sample).

## 3 Experiments

We evaluated the model on MNIST handwritten digits dataset. The unsupervised learning step was run for 5 epochs using the parameters: timestep $\Delta t = 0.01$, learning time constant $\tau_p = 60$, input-to-hidden connectivity $p_{i-h} = 8\%$ [5,6]. The input layer had $n_{HC}^{inp} \times n_{MC}^{inp} = 784x2$, hidden layer had $n_{HC}^{hid} \times n_{MC}^{hid} = \{30,100,200\}x100$, and the output layer had $n_{HC}^{out} \times n_{MC}^{out} = 1x10$. We ran the unsupervised learning step 5 times with random seeds. For semi-supervised learning, we generated training data by stratified random sampling (without replacement) $n = \{10, 20, 50, 100, 200, ... , 50000\}$ of the internal representations, ensuring all training sets have equal number of samples per class. For the validation set, we randomly sampled $n = 10000$ from the remaining samples. We repeated this 5 times with random seeds, so in total, for each network configuration we ran 25 random runs. Assoc. classifier was run for 5 epochs, and Go, No-go, and Go/No-go for 20 epochs each. The accuracy scores were reported for the validation set (mean ± s.d.). For comparative evaluation, we also used autoencoder (AE) and contractive autoencoder (CAE) networks, using the Adam optimizer with L2 reconstruction loss (minibatches of 256 samples for 300 epochs). For the CAE regularization we added the Frobenius norm (λ=0.01), and for AE - L1 sparsity loss (λ=1e-7). The linear classifier was trained using cross-entropy loss and Adam optimizer (minibatches of 256 samples for 300 epochs).

### 3.1 Evaluation of Go/No-go semi-supervised classifier

We first evaluated the Go, No-go, and Go/No-go network trained on the BCPNN internal representations. Fig. 3 shows the performance comparison for three different network sizes. We see that the Go and Go/No-go classifiers perform poorly in the low sample number regime, possibly because they quickly learn to select the correct labels and saturate prematurely. The No-go network, on the other hand, also learns to demote incorrect choices and hence, performs better. Go/No-go performs consistently better for all three network architectures over a wide range of samples and thus, we will use it further for our comparison.

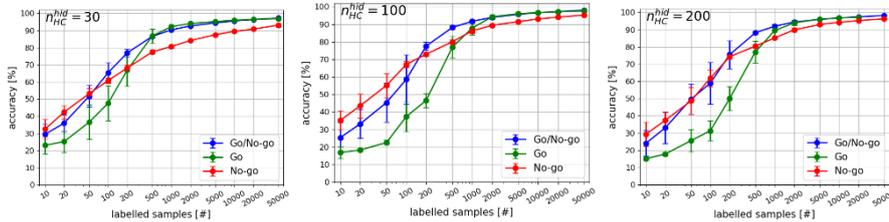

Fig. 3: Accuracy obtained with semi-supervised learning for Go, No-go, and Go/No-go trained on MNIST data with $n_{HC}^{hid} = 30$ (left), 100 (center), and 200 (right).

### 3.2 Comparison of BCPNN semi-supervised classifiers

Next, we compared different semi-supervised classifiers. Fig. 4 shows simple associative classifier performs consistently better than Go/No-go and linear classifier data (by 2% to 10%, p<0.01, Kruskal-Wallis test, N=25) when the number of labelled samples is small (<5000). For large data regimes however, Go/No-go and linear classifiers reach better performance, while the accuracy of the associative classifier flattens. This illustrates the sharp contrast between associative classifier when compared with other classifiers, and the specialization of different classifiers depending on data regimes. With few samples to train from, both the error-based learning methods (Go/No-go and linear classifiers) saturate quickly by predicting the right labels and drive the error towards zero, while the associative method keeps learning.

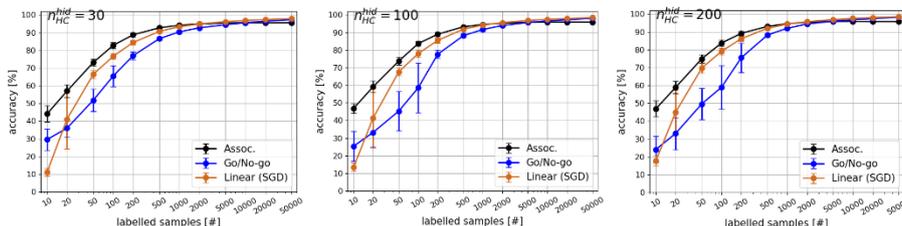

Fig. 4: Accuracy obtained with semi-supervised learning for associative, Go/No-go, and linear classifiers with $n_{HC}^{hid} = 30$ (left), 100 (center), and 200 (right).

### 3.3 Comparison of BCPNN with other models

Given that the BCPNN generated internal representations performed relatively well in semi-supervised settings when coupled with the Assoc. classifier, we evaluated the model with other popular semi-supervised learning baselines. We used the network with $n_{HC}^{hid} = 200$. Table 1 shows that BCPNN coupled with Assoc. classifier outperforms AE and CAE models until 1000 samples (p<0.01, Kruskal-Wallis test, N=25), and classifies the validation set at the level of ~47% with just one training sample per label. Considering these are classification done on internal representations generated from fully unsupervised learning, the associative classifier generalizes very well. For large data regimes however, the linear classifier, and the Go/No-go classifier to a slightly lesser extent, achieve much better performance (compared to baselines).

| Models | 10 | 100 | 1000 | 10000 | 50000 |
|---|---|---|---|---|---|
| BCPNN + Assoc. | **47.0±4.4** | **83.8±1.7** | **94.6±0.5** | 96.2±0.3 | 96.1±0.3 |
| BCPNN + Go/No-go | 34.9±7.6 | 72.8±8.6 | 93.2±0.5 | 96.8±0.2 | 97.5±0.0 |
| BCPNN + Linear (SGD) | 17.8±2.7 | 79.1±2.0 | **94.4±0.3** | **97.6±0.1** | **98.5±0.1** |
| AE + Linear (SGD) | 26.8±8.1 | 73.8±5.4 | 88.3±3.9 | 96.7±0.3 | 97.4±0.2 |
| CAE + Linear (SGD) | 33.1±5.3 | 74.6±3.3 | 91.1±0.1 | 98.3±0.1 | 98.2±0.0 |
| CAE + backprop finetuned [4] | - | 86.5 | 95.23 | - | 98.8 |
| Semi-sup. Embedding [2] | - | 83.1 | 94.3 | - | 98.5 |
| Transductive SVM [2] | - | 83.3 | 94.6 | - | 98.6 |
| MLP, BN, Gaussian noise [2] | - | 78.3±1.8 | 94.3±0.2 | - | 99.2±0.1 |

Table 1: Classification accuracies for different numbers of labelled samples. Peak BCPNN accuracies are written in bold (p<0.01, Kruskal-Wallis test, N=25).

## 4  Summary

We demonstrated that the internal representations generated in an unsupervised manner with a brain-like neural network architecture can be used in semi-supervised learning scenarios. For this, we introduced two classifiers, associative and Go/No-go, both of which are based on biologically plausible mechanisms. When compared on MNIST data with other semi-supervised baseline models, the associative classifier excels in the low data regime (until 1000 labeled samples), while the Go/No-go and linear classifiers perform better when trained on large number of labeled samples. One possibility is to combine associative and Go/No-go classifiers, gaining the advantages of each in different data regimes.